\documentclass[conference]{IEEEtran}
\usepackage[utf8]{inputenc}
\usepackage{graphicx}
\usepackage{amsmath,amssymb,amsfonts}
\usepackage[hidelinks]{hyperref}
\usepackage{cite}
\usepackage{subcaption}
\usepackage{caption}
\usepackage{array}
\newcommand{\bk}{\discretionary{}{}{}}

\title{NPU Offloading of a Frozen Visual Encoder for Robot Policy Training}

\author{
\IEEEauthorblockN{Hyojun Yun$^{1}$, Seungjae Won$^{2}$, Hyungpil Moon$^{1,2,\dagger}$}
\IEEEauthorblockA{
$^{1}$Department of Mechanical Engineering, $^{2}$Department of Intelligent Robotics \\
Sungkyunkwan University, Suwon, Korea \\
yhj0971@g.skku.edu, wonsj@g.skku.edu, $^{\dagger}$hyungpil@g.skku.edu
}
}

\begin{document}
\maketitle

\begin{abstract}
When a robot policy is trained for a new task or dataset, its visual encoder can be frozen and only its action generation module can be trained to reduce training cost.
Freezing the visual encoder removes its backward pass.
However, its forward pass must still run at every training step because the input images change.
This repeated computation continues to use GPU compute resources.
It is therefore important to determine whether moving this computation to a low power AI accelerator, such as an NPU, can reduce total energy despite the added data transfer and longer training time.
Its effect on policy performance must also be evaluated.
To study these questions, we built an asynchronous training pipeline that uses both a GPU and an NPU for the AR-Actor specialist.
The frozen visual encoder runs in A8W8 INT8 on a Mobilint Aries2 NPU, while the FP32 action expert is trained on an NVIDIA GeForce RTX 5060 Ti GPU.
We compared a GPU-only baseline with four conditions, L1 to L4.
These conditions gradually extend NPU offloading from one to four Transformer encoder layers.
Each condition was trained for 30,000 steps with three random seeds.
For the GPU-only condition, we measured GPU board power.
For the NPU conditions, we added GPU and NPU board power.
Energy per sample decreased by 17.1\% in L1, which offloaded ResNet18 and the first encoder layer.
It decreased by 27.9\% in L4, which offloaded ResNet18 and all four encoder layers.
In contrast, training time per sample increased by 15.2\% in L1 and 37.7\% in L4.
Peak allocated GPU memory decreased by 19.8 to 20.7\% in the NPU conditions.
The five conditions were trained with three training seeds, producing 15 policies.
Each policy was evaluated with the same 300 environment seeds, for a total of 4,500 simulator rollouts.
The combined success rate was 93.33\% for GPU-only and 91.44 to 92.89\% for the NPU conditions.
These results show that NPU offloading of a frozen visual encoder can reduce training energy, but it increases training time and lowers policy success rate by 0.44 to 1.89 percentage points compared with GPU-only training.
\end{abstract}


\section{Introduction}
\label{sec:intro}

Visual robot policies generate actions from camera observations.
To use such a policy for a new task or robot environment, the policy must be trained on data from that setting.
Recent vision language action policies use large vision language models.
As these models grow, they require more power and GPU memory during training.

Prior work has addressed this problem by freezing a pretrained visual encoder and training only the action generation module.
SmolVLA focuses on the high training cost of existing VLAs and the difficulty of deploying them.
It proposes a lightweight model that combines a small VLM with an action expert.
Its VLM is frozen, so only the action expert is trained.
The model is also designed to support training on a single GPU \cite{smolvla}.
This form of partial training does not need gradients or optimizer states for the frozen part.
It can therefore reduce the power and GPU memory required to train the full model.

However, freezing parameters does not remove all computation in the frozen part.
At every training step, the visual input changes.
The frozen visual encoder must therefore generate new features for the action expert, and this forward pass still runs on the GPU.
Prior work on lightweight robot policies mainly focuses on reducing model size and overall training or inference cost.
The effect of moving the repeated forward pass of a frozen visual encoder away from the GPU has not been studied in detail.

Some studies have moved the computation of a frozen model to a separate accelerator.
FixyNN freezes the early layers of a MobileNet pretrained on ImageNet and trains the remaining layers on new image classification datasets \cite{fixynn}.
The frozen early layers run on a fixed weight feature extractor, while the back end for each dataset runs on a programmable CNN accelerator.
This design provides high energy efficiency during inference.
However, its energy results are based on ASIC hardware modeling and simulation for a TSMC 16 nm process.
It does not measure training energy on a physical chip or evaluate robot policy training.

In our previous work \cite{npukroc}, we ran the first 14 of Qwen2-VL's 28 Transformer blocks on an NPU.
The NPU sent one LayerNorm1 output at the split point to the GPU.
The GPU combined this feature with action tokens.
A separate action prediction module with 12 Transformer blocks then processed these tokens with self attention.
This design did not use the K/V values from each backbone layer in the action expert.
Instead, it processed one intermediate backbone activation together with the action tokens.
The experiment also used a batch size of 1 and only 10,000 training steps.
It did not evaluate policy success on a specific robot task.

To address these limits, we chose the AR-Actor specialist from AR-VLA as the target model.
AR-VLA proposes an autoregressive action expert that keeps action history and uses it to generate the next action.
It defines a generalist that uses a VLM and a specialist called AR-Actor that uses ResNet18 \cite{arvla}.
The AR-Actor action expert uses the K/V values from all four encoder layers.
Unlike our previous work, which passed only one intermediate activation, the present work keeps the K/V values from all encoder layers after splitting the model between the GPU and NPU.
The NPU runs the forward pass of the frozen visual encoder, which does not require weight gradients.
The trainable action expert and optimizer remain on the GPU.
The GPU uses intermediate visual feature tokens sent from the NPU to reconstruct the K/V values for each layer.
We gradually change the offloading boundary and measure energy, training time, GPU memory, and policy success rate together.

The main contributions of this work are as follows.

\begin{enumerate}
\item We implemented an asynchronous training pipeline that uses a GPU and an NPU. It runs the frozen visual encoder on a Mobilint Aries2 NPU and trains the action expert on an RTX 5060 Ti GPU.
\item We defined four offloading boundaries. All boundaries include ResNet18, followed by the first one (L1), two (L2), three (L3), or four (L4) Transformer encoder layers. We compared these conditions with a GPU-only baseline using three training seeds. We measured power, energy, time, and GPU memory. We also evaluated the resulting 15 checkpoints with the same 300 environment seeds, for a total of 4,500 simulator rollouts.
\item We measured the transfer of intermediate visual feature tokens between the NPU and GPU, K/V reconstruction on the GPU, and the NPU wait time that remains after asynchronous overlap. These measurements show the limits of the current implementation.
\end{enumerate}

The rest of this paper is organized as follows.
Section~\ref{sec:method} describes the target policy, the two training phases, the NPU offloading boundaries, and asynchronous execution.
Section~\ref{sec:setup} presents the hardware, training data, comparison conditions, and measurement methods.
Section~\ref{sec:results} reports the results for energy, training time, GPU memory, and policy success rate.
Section~\ref{sec:discussion} discusses the meaning of the results and the limitations of this study.
Section~\ref{sec:conclusion} concludes the paper.

\section{Method}
\label{sec:method}

\subsection{Target Policy and Training Scope}

The AR-Actor specialist uses a ResNet18 visual backbone, an image projection module, and a four-layer Transformer encoder to extract visual features \cite{arvla}.
In this paper, we refer to these three parts together as the visual encoder.
We first train the full policy.
We then freeze the visual encoder, initialize a new action expert, and train it.
Our implementation has 51,515,228 parameters.
Table~\ref{tab:params} shows the model split.

\begin{table}[tb]
\caption{PARAMETER COUNT AND TRAINING STATE OF EACH POLICY MODULE}
\label{tab:params}
\centering
\footnotesize
\setlength{\tabcolsep}{4pt}
\hyphenpenalty=10000
\exhyphenpenalty=10000
\begin{tabular}{>{\raggedright\arraybackslash}p{1.45cm}>{\raggedright\arraybackslash}p{2.75cm}>{\raggedright\arraybackslash}p{1.75cm}>{\raggedright\arraybackslash}p{1.55cm}}
\hline
Part & Components & Number of parameters & State \\
\hline
Visual encoder & ResNet18, image projection, four Transformer encoder layers & 28,771,520 & Frozen \\
\hline
Action expert & State projection, four autoregressive Transformer decoder layers, one parallel decoder layer, and action head & 22,743,708 & Newly initialized and trained \\
\hline
\end{tabular}
\end{table}

\subsection{Two Training Phases}

The experiment has two phases.
Phase A prepares the visual encoder used in the comparison.
Phase B compares GPU-only training with four NPU offloading conditions while the visual encoder is frozen.
Both phases use a bimanual ALOHA task in which the robot transfers a cube from one hand to the other.
The training data contain 50 demonstration episodes and 20,000 frames.

In Phase A, we trained the full policy on the GPU for 200,000 steps.
We then evaluated the policy for 10 episodes in the ALOHA cube transfer simulator, and it succeeded in all 10 episodes.
We froze the visual encoder parameters from this checkpoint and used the same parameters in every comparison condition.

In Phase B, we combined the frozen visual encoder with a newly initialized action expert.
We trained each condition for 30,000 steps.
We initialized a new action expert so that all five conditions could be compared from the same initial state and with the same training setup.

Within each training seed, the five conditions used the same initial action expert weights, data order, random history masks, optimizer settings, and number of training steps.
We used AdamW with a learning rate of $1\times10^{-5}$, a weight decay of $1\times10^{-4}$, and a batch size of 8.

\subsection{Offloading Boundaries}

In the GPU-only condition, the entire frozen visual encoder runs in FP32 on the GPU.
In the NPU conditions, ResNet18, the image projection module, and the first one to four encoder layers run in A8W8 INT8 on the NPU.
Table~\ref{tab:conditions} summarizes the five comparison conditions.

\begin{table*}[t]
\caption{OFFLOADING BOUNDARIES AND THE COMPUTATION THAT REMAINS ON THE GPU}
\label{tab:conditions}
\centering
\begin{tabular}{llll}
\hline
Condition & Runs on NPU & Frozen computation remaining on GPU & Trained on GPU \\
\hline
GPU-only & None & Entire visual encoder & Action expert \\
L1 & ResNet18 + projection + encoder 0 & Encoders 1--3 + K/V reconstruction & Action expert \\
L2 & L1 range + encoder 1 & Encoders 2--3 + K/V reconstruction & Action expert \\
L3 & L2 range + encoder 2 & Encoder 3 + K/V reconstruction & Action expert \\
L4 & L3 range + encoder 3 & K/V reconstruction + final normalization & Action expert \\
\hline
\end{tabular}
\end{table*}

The AR decoder uses the keys and values from all four encoder layers.
The NPU therefore sends the Transformer encoder input tokens and the output tokens of each encoder layer that runs on the NPU to the GPU.
The GPU then reconstructs the K/V values for each encoder layer.
We first tried to make the NPU output the K/V values from each layer directly, but the NPU compiler could not compile this graph.
In L1, the NPU sends the Transformer encoder input tokens and the output tokens of the first encoder layer.
In L4, it also sends the output tokens of the second, third, and fourth layers.
One token array is 4.92 MB for a batch size of 8.
Thus, the transfer size per batch increases from 9.8 MB for two arrays in L1 to 24.6 MB for five arrays in L4.

\subsection{Asynchronous Overlap}

\begin{figure*}[t]
\centering
\includegraphics[width=0.92\textwidth]{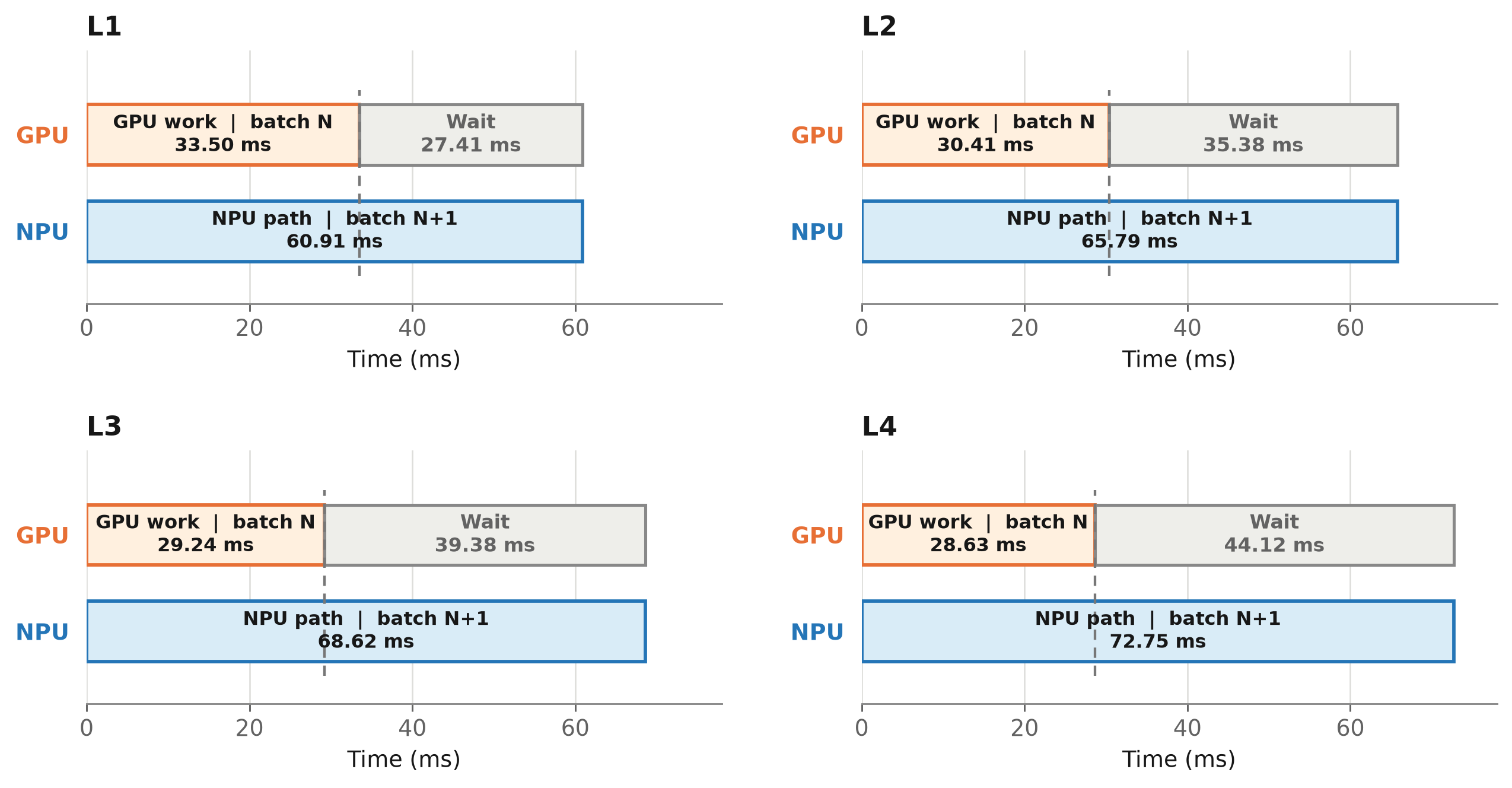}
\caption{Time breakdown of asynchronous GPU and NPU overlap by condition. While the GPU processes batch N, the NPU runs the frozen visual encoder forward pass for batch N+1 in advance.}
\label{fig:overview}
\end{figure*}

While the GPU trains the action expert for batch N, the NPU computes the frozen visual encoder for batch N+1 in advance.
A single background thread runs the NPU forward pass for the next batch.
The NPU path is longer than the GPU training path.
As a result, some GPU wait time remains even after the two operations overlap.
Figure~\ref{fig:overview} shows the time breakdown of this asynchronous process for each condition.

\section{Experimental Setup}
\label{sec:setup}

\subsection{Hardware, Software, and Quantization}

Table~\ref{tab:setup} lists the hardware, software, and precision settings used in all experiments.

\begin{table}[tb]
\caption{HARDWARE, SOFTWARE, AND PRECISION SETTINGS}
\label{tab:setup}
\centering
\footnotesize
\begin{tabular}{p{1.9cm}p{5.7cm}}
\hline
Item & Setting \\
\hline
CPU & Intel Core i5-10400 @ 2.90 GHz \\
GPU & NVIDIA GeForce RTX 5060 Ti 16 GB, driver 580.173.02 \\
NPU & Mobilint Aries2 16 GB, driver 1.12.0, firmware 1.2.5 \\
Device connection & The GPU and NPU are each connected to the host through PCIe \\
NPU tools & qbcompiler 1.2.0, qbruntime 1.2.0 \\
Training software & PyTorch 2.9.1+cu130, LeRobot 0.3.4, Python 3.10.19 \\
Policy code & \texttt{insait-institute/\bk AR-VLA-lerobot}, commit \texttt{e34012a9} \\
Precision & FP32 GPU training, A8W8 INT8 NPU visual encoder \\
\hline
\end{tabular}
\end{table}

For NPU quantization calibration, we used 48 real frames sampled evenly from episodes 0 to 7 of the training data.
We measured the match between NPU and GPU outputs to check whether the visual features remained close to the GPU FP32 output after INT8 quantization and NPU execution.
For this test, we used 40 frames from episodes 40 to 49, which were not used for calibration.
At each split boundary between the NPU and GPU, we compared the visual feature tokens from the INT8 NPU path with the GPU FP32 output at the same boundary.
The cosine similarities were 0.99752 for L1, 0.99763 for L2, 0.99782 for L3, and 0.99783 for L4.

\subsection{Data and Evaluation Environment}

The training dataset was \texttt{lerobot/\bk aloha\_\bk sim\_\bk transfer\_\bk cube\_\bk scripted}, which contains 50 episodes and 20,000 frames.
The input included one overhead RGB camera and a robot state with 14 dimensions.
The output was an action with 14 dimensions.
Images were normalized with the ImageNet mean and standard deviation.
States and actions were normalized with the dataset statistics.
The bimanual ALOHA environment is based on Zhao et al. \cite{aloha}.
We ran the evaluation in the \texttt{AlohaTransferCube-v0} simulator environment.

We evaluated policy performance with continuous control rollouts from 300 initial conditions generated by environment seeds 2000 to 2299.

\subsection{Comparison Conditions and Repetitions}

We ran each of the five conditions, GPU-only and L1 to L4, once for each action expert training seed: 2000, 2001, and 2002.
This produced 15 checkpoints.
Before each run, we confirmed that no other process was using the GPU for computation and that the NPU was idle.
We allowed 20 seconds for the system to stabilize between runs.

We evaluated every checkpoint with the same 300 environment seeds.
The GPU-only policy ran the full visual encoder on the GPU.
Each NPU policy used the same NPU precision and split boundary from L1 to L4 during training and evaluation.
The total number of rollouts was $15\times300=4{,}500$.

\subsection{Power and Energy}

We measured GPU board power with the \texttt{nvidia-smi} command and NPU board power with the \texttt{mobilint-cli status} command.
Power from both devices was recorded once per second.
For each run, we integrated the recorded power from the start to the end of training with the trapezoidal rule.
We divided this value by the number of processed samples to calculate energy per sample.

\begin{equation}
E_{\mathrm{sample}}=\frac{1}{N}\int_{t_0}^{t_1}\sum_{d\in\mathcal{A}}P_d(t)\,dt,
\end{equation}

Here, $N$ is the number of processed training samples, and $\mathcal{A}$ is the set of accelerator boards used in a condition.
For GPU-only, $\mathcal{A}=\{\mathrm{GPU}\}$.
For the NPU conditions, $\mathcal{A}=\{\mathrm{GPU},\mathrm{NPU}\}$.
Thus, the reported energy is not total system energy.
It is the combined energy calculated from the GPU and NPU board power reported by the commands above.
It does not include the energy or loss from the CPU, DRAM, PCIe, or power supply.

\subsection{GPU Memory}

We measured GPU memory as the peak amount of memory that PyTorch allocated to tensors during training.
Training feasibility and batch size depend on the point of highest memory use.
We therefore used the peak value instead of the average.

\section{Results}
\label{sec:results}

\subsection{Accelerator Board Energy and Training Time}

\begin{table*}[t]
\caption{MEAN BOARD POWER, ENERGY, AND TRAINING TIME}
\label{tab:energy}
\centering
\begin{tabular}{lrrrrrrrr}
\hline
 & Mean GPU & Mean NPU & Mean combined & Total training & Energy per & vs. & Time per & Throughput \\
Condition & board power & board power & board power & time & sample & GPU-only & sample & \\
 & (W) & (W) & (W) & (min) & (J) & (\%) & (ms) & (sample/s) \\
\hline
GPU-only & 138.59 & --- & 138.59 & 26.43 & 0.91512 & --- & 6.607 & 151.4 \\
L1 & 80.85 & 18.81 & 99.66 & 30.46 & 0.75831 & $-17.1$ & 7.613 & 131.3 \\
L2 & 70.67 & 18.88 & 89.55 & 32.94 & 0.73671 & $-19.5$ & 8.224 & 121.6 \\
L3 & 62.84 & 19.08 & 81.92 & 34.31 & 0.70244 & $-23.2$ & 8.578 & 116.6 \\
L4 & 53.35 & 19.19 & 72.54 & 36.38 & 0.65955 & $-27.9$ & 9.094 & 110.0 \\
\hline
\end{tabular}
\\[2pt]
\parbox{\textwidth}{\footnotesize The mean board power of each run is the average of the values recorded once per second during training. The table reports the mean across three training seeds. Total training time is the time required for 30,000 steps with a batch size of 8. For the NPU conditions, mean combined board power is the sum of the mean GPU and NPU board power used at the same time.}
\end{table*}

\begin{figure}[tb]
\centering
\includegraphics[width=\columnwidth]{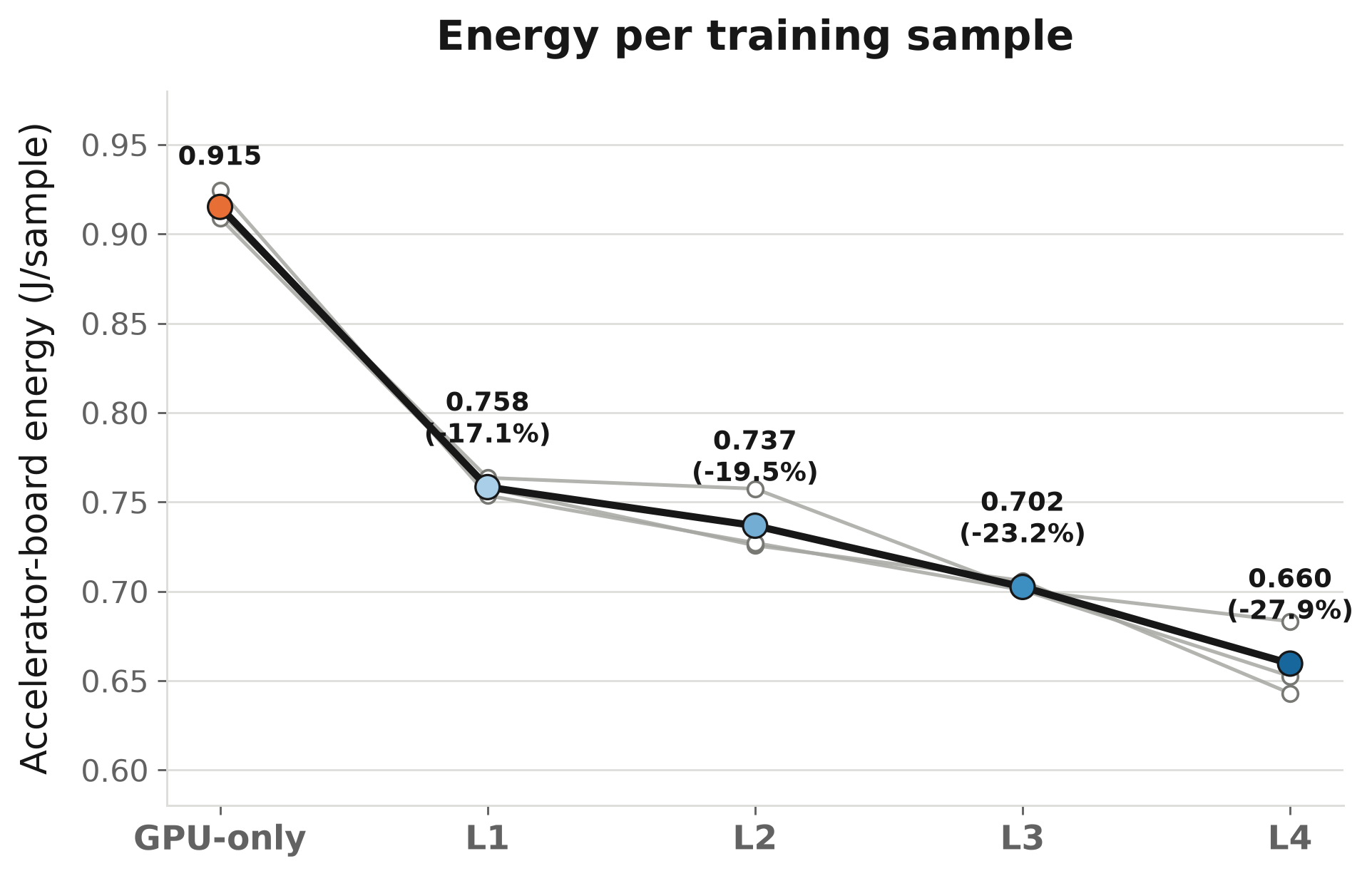}
\caption{Accelerator board energy per sample by offloading boundary. Gray lines connect the conditions measured with the same training seed. Colored points show the mean across the three training seeds.}
\label{fig:energy}
\end{figure}

Table~\ref{tab:energy} reports the mean board power, energy, and training time of each condition.
As shown in Figure~\ref{fig:energy}, all four NPU boundaries used less energy per sample than GPU-only.
Energy also decreased from L1 to L4 for each of the three training seeds.

Time per sample in L4 was 37.7\% longer than in GPU-only, but combined board power was 47.7\% lower.
The power reduction was larger than the time increase, so energy per sample was 27.9\% lower.

\subsection{GPU Memory}

\begin{figure}[tb]
\centering
\includegraphics[width=\columnwidth]{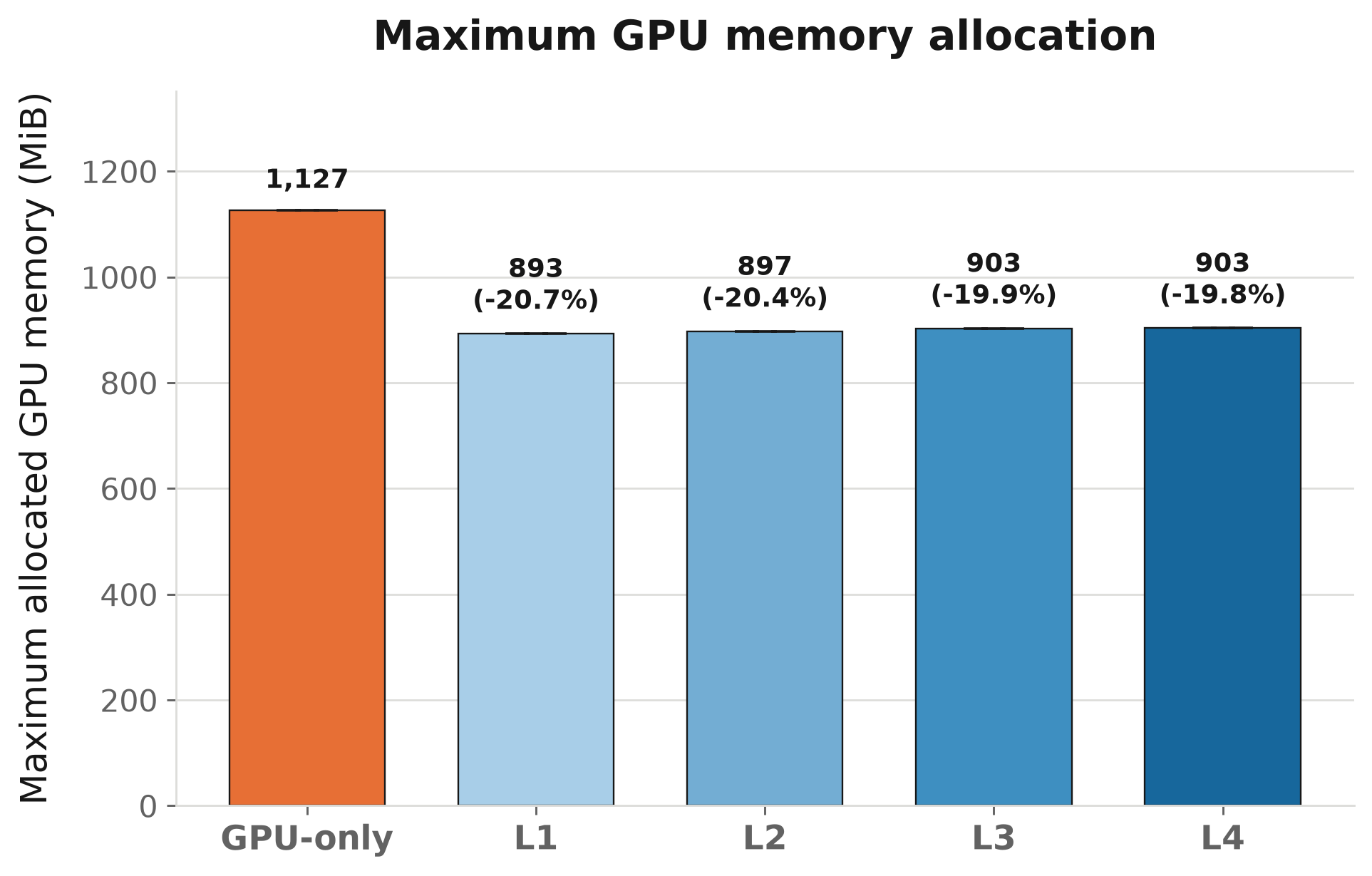}
\caption{Peak allocated GPU memory by condition. Each bar shows the mean across three training seeds.}
\label{fig:memory}
\end{figure}

Figure~\ref{fig:memory} shows the peak allocated GPU memory for each condition.
Peak memory decreased by 234 MiB from GPU-only to L1, but changed little from L1 to L4.
The AR decoder uses K/V values from all four encoder layers.
Therefore, the normalization and K/V projection weights needed to reconstruct each layer's K/V values always remain on the GPU, regardless of the offloading boundary.
In our implementation, the NPU compiler could not compile a graph that directly outputs the K/V values from each layer.
We therefore sent intermediate visual feature tokens from the NPU to the GPU and reconstructed the K/V values on the GPU.
A deeper boundary reduces the encoder activations computed on the GPU, but it increases the number of token arrays sent by the NPU from two to five.
Because of these two effects, peak allocated memory does not continue to decrease as the boundary becomes deeper.

\subsection{Simulator Policy Performance}

\begin{figure}[tb]
\centering
\includegraphics[width=\columnwidth]{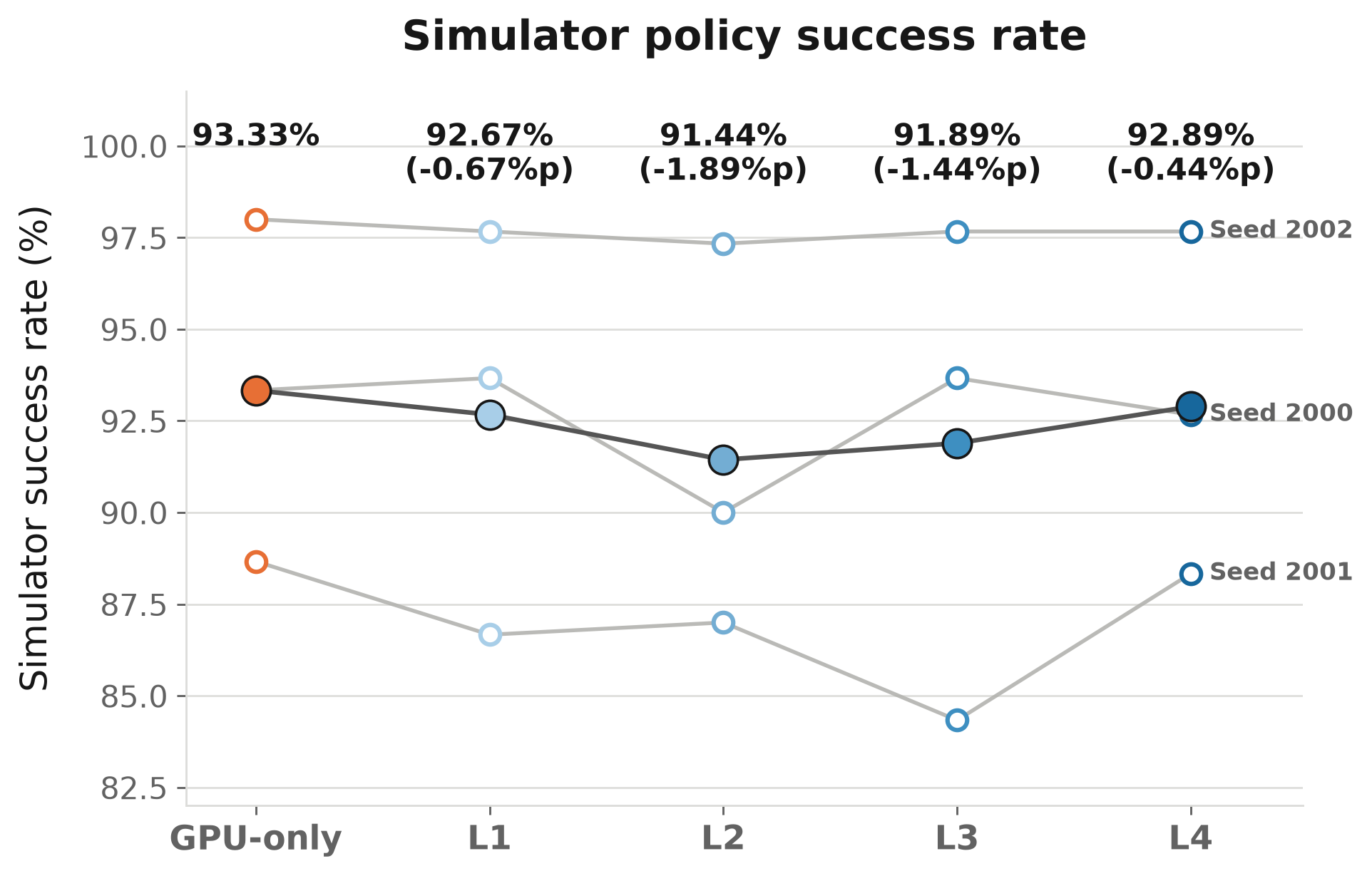}
\caption{Simulator policy success rate by condition. Large colored points and the dark line show the success rate over 900 evaluations for each condition. Small open points and gray lines connect checkpoint results from the same training seed. The training seed is shown at the right end of each gray line. Values in parentheses show the difference from GPU-only.}
\label{fig:quality}
\end{figure}

Each condition was trained with three training seeds, producing three checkpoints per condition.
Each checkpoint was evaluated with the same 300 environment seeds.
Thus, each condition had 900 evaluations in total.
Figure~\ref{fig:quality} shows the success rate for each condition and the result from each training seed.
The success rate was 93.33\% for GPU-only and 91.44 to 92.89\% for the NPU conditions.
The NPU success rates were 0.44 to 1.89 percentage points lower than the GPU-only result.

\section{Discussion and Limitations}
\label{sec:discussion}

\subsection{Meaning of the Energy Reduction}

In our experiments, NPU offloading reduced throughput but also reduced the accelerator energy needed per training sample.
The deepest boundary, L4, was 37.7\% slower than GPU-only.
However, its combined board power was 47.7\% lower, so its energy per sample was 27.9\% lower.

This result cannot be explained by NPU hardware alone.
The GPU-only and NPU conditions differ not only in the devices that run the model.
Data movement among the GPU, CPU, and NPU also changes.
The scheduling method also changes.
Our results directly show only that the implemented system reduced total accelerator board energy.

\subsection{Limitations}

First, we evaluated only one AR-Actor specialist, one ALOHA simulation task, and one system with a GPU and an NPU.
More experiments are needed to test other policy structures, dataset sizes, batch sizes, and device combinations.

Second, the device, numerical precision, data transfer, and scheduling all change at the same time.
Therefore, this study does not compare GPU and NPU hardware efficiency as a single isolated factor.

Third, we measured only the accelerator board power reported by \texttt{nvidia-smi} and \texttt{mobilint-cli}.
We did not include the CPU, DRAM, PCIe, or power supply losses.
The reduction in total system energy may therefore differ from our results.

Fourth, each condition was run once for each of three different training seeds.
Because the number of runs is small, we cannot determine whether small performance differences between conditions would remain consistent in repeated experiments.

Fifth, the compiler failed to compile the graph that directly outputs K/V values.
We could therefore not place the entire visual encoder on the NPU.
Even in L4, the computation and tensors needed for K/V reconstruction remain on the GPU.

\section{Conclusion}
\label{sec:conclusion}

We implemented an asynchronous training pipeline that runs the frozen visual encoder of the AR-Actor specialist on an energy efficient INT8 NPU and trains its FP32 action expert on a GPU.
Across three training seeds, NPU offloading reduced energy per sample on the active accelerator boards by 17.1 to 27.9\%.
In return, it increased training time per sample by 15.2 to 37.7\%.
Peak allocated GPU memory during training decreased by 19.8 to 20.7\% in the current implementation.

The policy success rate was 93.33\% for GPU-only and 91.44 to 92.89\% for the NPU conditions.
However, with only three training seeds, we cannot determine the exact size of the performance decrease.
The main result of this study is that NPU offloading of a frozen visual encoder can reduce the total accelerator board energy used for robot policy training in a working system that uses both a GPU and an NPU.

Future work should focus on three directions.
First, the NPU should directly output the K/V values from each encoder layer.
This would reduce the transfer of intermediate visual feature tokens and the K/V reconstruction work on the GPU.
Second, total system power should be measured with an external precision power meter instead of using values reported by device commands.
Third, the proposed offloading method should be tested with different GPU and NPU combinations, robot environments and tasks, and robot policy models.

\section*{Acknowledgment}

This work was supported by the Robot Industrial Technology Development Program (RS-2024-00444054, Development of low-cost, high-performance visual sensor technology integrated with AI semiconductors and AI algorithms designed for robots) funded by the Ministry of Trade, Industry and Energy (MOTIE, Korea).

\end{document}